\documentclass[final]{cvpr}

\usepackage{times}
\usepackage{epsfig}
\usepackage{graphicx}
\usepackage{multirow}
\usepackage{amsmath}
\usepackage{amssymb}
\usepackage{booktabs}
\usepackage{draftwatermark}
\usepackage[pagebackref=true,breaklinks=true,colorlinks,bookmarks=false]{hyperref}

\def\cvprPaperID{19} 
\def\confYear{CVPR 2021}

\begin{document}

\title{Evaluating Deep Neural Networks Trained on Clinical Images in Dermatology with the Fitzpatrick 17k Dataset}

\author{Matthew Groh\\
MIT Media Lab\\
Cambridge, MA\\
{\tt\small groh@mit.edu}
\and
Caleb Harris\\
MIT Media Lab\\
Cambridge, MA\\
{\tt\small harrisc@mit.edu}
\and
Luis Soenksen\\
MIT, Harvard University\\
Cambridge, MA\\
{\tt\small soenksen@mit.edu}
\and
Felix Lau\\
Scale\\
San Francisco, CA\\
{\tt\small felix.lau@scale.com}
\and
Rachel Han\\
Scale\\
San Francisco, CA\\
{\tt\small rachel.han@scale.com}
\and
Aerin Kim\\
Scale\\
San Francisco, CA\\
{\tt\small aerin.kim@scale.com}
\and
Arash Koochek\\
Banner Health\\
Phoenix, AZ\\
{\tt\small arash.koochek@bannerhealth.com}
\and
Omar Badri\\
Northeast Dermatology Associates\\
Beverly, MA\\
{\tt\small obadri@gmail.com}
}

\maketitle

\begin{abstract}

How does the accuracy of deep neural network models trained to classify clinical images of skin conditions vary across skin color? While recent studies demonstrate computer vision models can serve as a useful decision support tool in healthcare and provide dermatologist-level classification on a number of specific tasks, darker skin is underrepresented in the data. Most publicly available data sets do not include Fitzpatrick skin type labels. We annotate 16,577 clinical images sourced from two dermatology atlases with Fitzpatrick skin type labels and open-source these annotations. Based on these labels, we find that there are significantly more images of light skin types than dark skin types in this dataset. We train a deep neural network model to classify 114 skin conditions and find that the model is most accurate on skin types similar to those it was trained on. In addition, we evaluate how an algorithmic approach to identifying skin tones, individual typology angle, compares with Fitzpatrick skin type labels annotated by a team of human labelers. 

\end{abstract}

\section{Motivation}

How does the accuracy of deep neural network models trained to classify clinical images of skin conditions vary across skin color? The emergence of deep neural network models that can accurately classify images of skin conditions presents an opportunity to improve dermatology and healthcare at large \cite{esteva_dermatologist-level_2017, liu_deep_2020, tschandl_humancomputer_2020, soenksen2021using, celebi2019dermoscopy}. But, the data upon which these models are trained are mostly made up of images of people with light skin. In the United States, dark skin is underrepresented in dermatology residency programs~\cite{lester_diversity_nodate}, textbooks~\cite{alvarado_representation_2020,adelekun2021skin}, dermatology research~\cite{lester_absence_2020}, and dermatology diagnoses~\cite{palmieri_missed_2019, gupta_skin_nodate}. With the exception of PAD-UFES-20~\cite{pacheco2020pad}, none of the publicly available data sets identified by the Sixth ISIC Skin Image Analysis Workshop at CVPR 2021 (Derm7pt~\cite{kawahara2018seven}, Dermofit Image Library, ISIC 2018 \cite{codella2018skin,tschandl2018ham10000}, ISIC 2019 \cite{codella2019skin,tschandl2018ham10000, combalia2019bcn20000}, ISIC 2020\cite{rotemberg2021patient, international2020siim}, MED-NODE \cite{giotis2015med}, PH2~\cite{mendoncca2013ph}, SD-128~\cite{sun2016benchmark}, SD-198, SD-260) include skin type or skin color labels or any other information related to race and ethnicity. The only dataset with such skin type labels, PAD-UFES-20, contains Fitzpatrick skin type labels for 579 out of 1,373 patients in the dataset. The lack of consideration of subgroups within a population has been shown to lead deep neural networks to produce large accuracy disparities across gender and skin color for facial recognition~\cite{buolamwini2018gender}, across images with and without surgical markings in dermatology \cite{winkler_association_2019, bissoto2020debiasing}, and across treated and untreated conditions in radiology~\cite{oakden-rayner_hidden_2020}. These inaccuracies arise from dataset biases, and these underlying data biases can unexpectedly lead to systematic bias against groups of people~\cite{barocas2016big,abbasi-sureshjani_risk_2020}. If these dataset biases are left unexamined in dermatology images, machine learning models have the potential to increase healthcare disparities in dermatology~\cite{adamson_machine_2018}. 

By creating transparency and explicitly identifying likely sources of bias, it is possible to develop machine learning models that are not only accurate but also serve as discrimination detectors \cite{obermeyer_dissecting_2019,kleinberg_algorithms_2020, cowgill2020algorithmic}. By rigorously examining potentials for discrimination across the entire pipeline for machine learning model development in healthcare~\cite{chen2020ethical}, we can identify opportunities to address discrimination such as collecting additional data from underrepresented groups~\cite{chen2018my} or disentangling the source of the disparities~\cite{pierson2021algorithmic}. In this paper, we present the \textit{Fitzpatrick 17k} dataset which is a collection of images from two online dermatology atlases annotated with Fitzpatrick skin types by a team of humans. We train a deep neural network to classify skin conditions solely from images, and we evaluate accuracy across skin types. 

We also use the \textit{Fitzpatrick 17k} dataset to compare Fitzpatrick skin type labels to a computational method for estimating skin tone: individual typology angle (ITA). ITA is promising because it can be computed directly from images, but its performance varies with lighting conditions and may not always be effective for accurately annotating clinical images with skin types ~\cite{10.1001/jamadermatol.2015.0351, Kolkur_2017,kinyanjui_estimating_2019}. 

\section{Fitzpatrick 17k Dataset}

The \textit{Fitzpatrick 17k} dataset contains 16,577 clinical images with skin condition labels and skin type labels based on the Fitzpatrick scoring system~\cite{fitzpatrick1988validity}. The dataset is accessible at \href{https://github.com/mattgroh/fitzpatrick17k}{https://github.com/mattgroh/fitzpatrick17k}.

The images are sourced from two online open-source dermatology atlases: 12,672 images from DermaAmin and  3,905 images from Atlas Dermatologico~\cite{alkattash, desilva}. These sources include images and their corresponding skin condition label. While these labels are not known to be confirmed by a biopsy, these images and their skin condition labels have been used and cited in dermatology and computer vision literature a number of times~\cite{esteva_dermatologist-level_2017,han2018classification, bissoto2018deep,ramezani2014automatic,arifin2012dermatological, swinney2015incontinentia, van2015cutaneous}. As a data quality check, we asked a board-certified dermatologist to evaluate the diagnostic accuracy of 3\% of the dataset. Based on a random sample of 504 images, a board-certified dermatologist identified 69.0\% of images as diagnostic of the labeled condition, 19.2\% of images as potentially diagnostic (not clearly diagnostic but not necessarily mislabeled, further testing would be required), 6.3\% as characteristic (resembling the appearance of such a condition but not clearly diagnostic), 3.4\% are considered wrongly labeled, and 2.0\% are labeled as other. A second board-certified dermatologist also examined this sample of images and confirmed the error rate. This error rate is consistent with the 3.4\% average error rate in the most commonly used test datasets for computer vision, natural language processing, and audio processing~\cite{northcutt2021pervasive}. 

We selected images to annotate based on the most common dermatology conditions across these two data sources excluding the following 22 categories of skin conditions: (1) viral diseases, HPV, herpes, molluscum, exanthems, and others (2) fungal infections, (3) bacterial infections, (4) acquired autoimmune bullous disease, (5) mycobacterial infection (6) benign vascular lesions (7) scarring alopecia, (8) non-scarring alopecia (9) keratoderma (10) ichthyosis, (11) vasculitis, (12) pellagra like eruption (13) reiters disease (14) epidermolysis bullosa pruriginosa (15) amyloidosis, (16) pernio and mimics (17) skin metastases of tumours of internal organs (18) erythrokeratodermia progressive symmetric, (19) epidermolytic hyperkeratosis, (20) infections, (21) generalized eruptive histiocytoma, (21) dry skin eczema. We excluded these categories because they were either too broad, the images were of poor quality, or the categories represented a rare genodermatosis. The final sample includes 114 conditions with at least 53 images (and a maximum of 653 images) per skin condition. 

This dataset also includes two additional aggregated levels of skin condition classification based on the skin lesion taxonomy developed by Esteva et al. 2017, which can be helpful to improve the explainability of a deep learning system in dermatology~\cite{esteva_dermatologist-level_2017, barata2021explainable}. At the highest level, skin conditions are split into three categories: 2,234 benign lesions, 2,263 malignant lesions, and 12,080 non-neoplastic lesions. At a slightly more granular level, images of skin conditions are split into nine categories: 10,886 images labeled inflammatory, 1,352 malignant epidermal, 1,194 genodermatoses, 1,067 benign dermal,  931 benign epidermal, 573 malignant melanoma, 236 benign melanocyte, 182 malignant cutaneous lymphoma, and 156 malignant dermal. At the most granular level, images are labeled by skin condition.

The images are annotated with Fitzpatrick skin type labels by a team of human annotators from Scale AI. The Fitzpatrick labeling system is a six-point scale originally developed for classifying sun reactivity of skin and adjusting clinical medicine according to skin phenotype~\cite{fitzpatrick1988validity}. Recently, the Fitzpatrick scale has been used in computer vision for evaluating algorithmic fairness and model accuracy across skin type~\cite{buolamwini2018gender,liu_deep_2020,dulmage_point--care_2020}. Fitzpatrick labels allow us to begin assessing algorithmic fairness, but we note that the Fitzpatrick scale does not capture the full diversity of skin types \cite{ware2020racial}. Each image is annotated with a Fitzpatrick skin type label by two to five annotators based on Scale AI's dynamic consensus process. The number of annotators per image is determined by a minimal threshold for agreement, which takes into account an annotator's historical accuracy evaluated against a gold standard dataset, which consists of 312 images with Fitzpatrick skin type annotations provided by a board-certified dermatologist. This annotation process resulted in 72,277 annotations in total.

In the \textit{Fitzpatrick 17k} dataset, there are significantly more images of light skin types than dark skin. There are 7,755 images of the lightest skin types (1 \& 2), 6,089 images of the middle skin types (3 \& 4), and 2,168 images of the darkest skin types (5 \& 6). Table~\ref{fig:T1} presents the distribution of images by skin type for each of the three highest level categorizations of skin conditions. A small portion of the dataset (565 images) are labeled as unknown, which indicates that the team of annotators could not reasonably identify the skin type within the image.

The imbalance of skin types across images is paired with an imbalance of skin types across skin condition labels. The \textit{Fitzpatrick 17k} dataset has at least one image of all 114 skin conditions for Fitzpatrick skin types 1 through 3. For the remaining Fitpatrick skin types, there are 113 skin conditions represented in type 4, 112 represented in type 5, and 89 represented in type 6. In other words, 25 of the 114 skin conditions in this dataset have no examples in Fitzparick type 6 skin. The mean Fitzpatrick skin types across these skin condition labels ranges from 1.77 for basal cell carcionma morpheaform to 4.25 for pityriasis rubra pilaris. Only 10 skin conditions have a mean Fitzpatrick skin type above 3.5, which is the expected mean for a balanced dataset across Fitzpatrick skin types. These 10 conditions include: pityriasis rubra pilaris, xeroderma pigmentosum, vitiligo, neurofibromatosis, lichen amyloidosis, confluent and reticulated papillomatosis, acanthosis nigricans, prurigo nodularis, lichen simplex, and erythema elevatum diutinum. 

\begin{table}[!htbp] \centering
\begin{tabular}{lrrr}
\toprule
                     & Non-Neoplastic   & Benign & Malignant \\
\midrule
\# Images               & 12,080           & 2,234    & 2,263                \\
\midrule
Type 1               & 17.0\%        & 19.9\%          &  20.2\%                                  \\
Type 2               & 28.1\%        & 30.0\%           & 32.8\%                         \\
Type 3               & 19.7\%        & 21.2\%           & 20.2\%                         \\
Type 4               & 17.5\%        & 16.4\%           & 13.3\%                         \\
Type 5               & 10.1\%        & 7.1\%           & 6.5\%                          \\
Type 6               & 4.4\%        & 2.0\%           & 2.7\%                          \\
Unknown              & 3.2\%       & 3.3\%           & 4.6\%                   \\
\bottomrule
\end{tabular}
\caption{Distribution of skin conditions in \textit{Fitzpatrick 17k} by Fitzpatrick skin type and high level skin condition categorization.}
\label{fig:T1}
\end{table}

\begin{table}[!htbp] \centering
\begin{tabular}{lrrr}
\toprule
                     & Accuracy   &  Accuracy (off-by-one) & \# of Images \\
\midrule
Type 1               & 49\%        & 79\%          &  10                                 \\
Type 2               & 38\%        & 84\%           & 100                        \\
Type 3               & 25\%        & 71\%           & 98                       \\
Type 4               & 26\%        & 71\%           & 47                         \\
Type 5               & 34\%        & 85\%           & 44                         \\
Type 6               & 59\%        & 83\%           & 13                         \\         
\bottomrule
\end{tabular}
\caption{Accuracy of human annotators relative to the gold standard dataset of 312 Fitzpatrick skin type annotations provided by a board-certified dermatologist.}
\label{fig:labeling}
\end{table}

\section{Classifying Skin Conditions with a Deep Neural Network}

\subsection{Methodology}

We train a transfer learning model based on a VGG-16 deep neural network architecture~\cite{simonyan2014very} pre-trained on ImageNet~\cite{deng2009imagenet}. We replace the last fully connected 1000 unit layer with the following sequence of layers: a fully connected 256 unit layer, a ReLU layer, dropout layer with a 40\% change of dropping, a layer with the number of predicted categories, and finally a softmax layer. As a result, the model has 135,335,076 total parameters of which 1,074,532 are trainable. We train the model by using the Adam optimization algorithm to minimize negative log likelihood loss. We address class imbalance by using a weighted random sampler where the weights are determined by each skin condition's inverse frequency in the dataset. We perform a number of transformations to images before training the model which include: randomly resizing images to 256 pixels by 256 pixels, randomly rotating images 0 to 15 degrees, randomly altering the brightness, contrast, saturation, and hue of each image, randomly flipping the image horizontally or not, center cropping the image to be 224 pixels by 224 pixels, and normalizing the image arrays by the ImageNet means and standard deviations.

We evaluate the classifier's performance via 5 approaches: (1) testing on the subset of images labeled by a board-certified dermatologist as diagnostic of the labeled condition and training on the rest of the data  (2) testing on a randomly selected 20\% of the images where the random selection was stratified on skin conditions and training on the rest of the data (3) testing on images from Atlas Dermatologico and training on images from Derma Amin (4) testing on images from Derma Amin and training on images from Atlas Dermatologico (5) training on images labeled as Fitzpatrick skin types 1-2 (or 3-4 or 5-6) and testing on the rest of the data. The accuracy on the validation set begins to flatten after 10 to 20 epochs for each validation fold. We trained the same architecture on each fold and report accuracy scores for the epoch with the lowest loss on the validation set.

\begin{table*}[ht]
\centering
\begin{tabular}{llllllll}
\toprule
Holdout Set & Verified & Random & Source A & Source B & Fitz 3-6 & Fitz 1-2 \& 5-6 & Fitz 1-4 \\
\midrule
\# Train Images & 16,229 &  12,751 & 12,672 & 3,905 & 7,755 & 6,089 & 2,168 \\
\# Test Images & 348 &  3,826 & 3,905 & 12,672 & 8,257 & 10,488 & 14,409 \\
\midrule
Overall      & 26.7\%        &    20.2\%       &   27.4\%  & 11.4\%     & 13.8\% & 13.4\% & 7.7\%                      \\
Type 1       & 15.1\%         &    15.8\%          & 40.1\%   & 6.6\%   & -     & 10.0\%     & 4.4\%                     \\
Type 2     & 23.9\%          &    16.9\%           & 27.7\%   & 8.6\%   & -     & 13.0\%     & 5.5\%                     \\
Type 3     & 27.9\%           &    22.2\%          & 25.3\%   & 13.7\%  & 15.9\%     & -    & 9.1\%                     \\
Type 4       & 30.9\%        &    24.1\%            & 26.2\%   & 17.1\% & 14.2\%     & -     & 12.9\%                     \\
Type 5       & 37.2\%        &    28.9\%          & 28.4\%   & 17.6\%   & 10.1\%     & 21.1\%     & -                     \\
Type 6     & 28.2\%          &    15.5\%      & 25.7\%   & 14.9\%       & 9.0\%     & 12.1\%     & -                      \\ 
\bottomrule
\end{tabular}
\caption{Accuracy rates classifying 114 skin conditions across skin types on six selections of holdout sets. The verified holdout set is a subset of a randomly sampled set of images verified by a board-certified dermatologist as diagnostic of the labeled condition. The random holdout set is a randomly sampled set of images. The source A holdout set are all images from Atlas Dermatologico. The source B holdout set are all images from Derma Amin. The 3 Fitzpatrick holdout sets are selected according to Fitzpatrick labels. In all cases, the training data are the remaining non-held out images from the \textit{Fitzpatrick 17k} dataset.}
\label{fig:T2}
\end{table*}

\begin{table}[!ht] \centering
\scalebox{0.80}{

\begin{tabular}{cc|c|c|c|}
 &\multicolumn{1}{c}{}&\multicolumn{3}{c}{\textbf{Predicted Class}}\\
&\multicolumn{1}{c}{}&\multicolumn{1}{c}{Benign}
&\multicolumn{1}{c}{Malignant}
&\multicolumn{1}{c}{Non-neoplastic}\\
\cline{3-5}
\multicolumn{1}{c}{\multirow{6}{*}{{\textbf{Actual Class}}}}
& {\multirow{2}{*}{{Benign}}} & {\multirow{2}{*}{{275}}} & {\multirow{2}{*}{{52}}} & {\multirow{2}{*}{{54}}} \\
& & & & 
\\
\cline{3-5}
& {\multirow{2}{*}{{Malignant}}} & {\multirow{2}{*}{{106}}} & {\multirow{2}{*}{{487}}} & {\multirow{2}{*}{{109}}} \\
&  &  & & \\
\cline{3-5}
& {\multirow{2}{*}{{Non-neoplastic}}} & {\multirow{2}{*}{{788}}} & {\multirow{2}{*}{{448}}} & {\multirow{2}{*}{{1586}}} \\
&  &  & & \\
\cline{3-5}

\end{tabular}
}
\caption{Confusion matrix for deep neural network performance on predicting the high-level skin condition categories in the holdout set of images from Atlas Dermatologico.}
\label{fig:confusion_0}
\end{table}

\subsection{Results}

We report results of training the model on all 114 skin conditions across 7 different selections of holdout sets in Table~\ref{fig:T2}.

In the random holdout, the model produces a 20.2\% overall accuracy on exactly identifying the labeled skin condition present in the image. The top-2 accuracy (the rate at which the first or second prediction of the model is the same as the image's label) is 29.0\% and the top-3 accuracy is 35.4\%. These numbers can be evaluated against random guessing, which would be 1/114 or 0.9\% accuracy. Across the 114 skin conditions, the median accuracy is 20.0\% and ranges from a minimum of 0\% accuracy on 10 conditions (433 images in the random holdout) and a maximum of 93.3\% accuracy on 1 condition (30 images). 

When we train the model on the 3 category partition of non-neoplastic, benign, and malignant, the model produces an accuracy of 62.4\% on the random holdout (random guessing would produce 33.3\% accuracy). Likewise, the model trained on the 9 category partition produces an accuracy of 36.1\% on the random holdout (random guessing would produce 11.1\% accuracy). Another benchmark for this 3 partition and 9 partition comes from Esteva et al. which trained a model on a dataset 7.5 times larger to produce 72.1\% accuracy on the 3 category task and 55.4\% accuracy on the 9 category task~\cite{esteva_dermatologist-level_2017}.

Depending on each holdout selection, the accuracy rates produced by the model vary across skin types. For the first four holdout selections in Table~\ref{fig:T2} – the verified selection, the random holdout, the source A holdout based on images from Atlas Dermatologico, and the source B holdout based on images from Derma Amin – we do not find a systematic pattern in accuracy scores across skin type. For the second three holdout selections where the model is trained on images from two Fitzpatrick types and evaluated on images in the other four Fitzpatrick types, we find the model is most accurate on the images with the closest Fitzpatrick skin types to the training images. Specifically, the model trained on images labeled as Fitzpatrick skin types 1 and 2 performed better on types 3 and 4 than types 5 and 6. Likewise, the model trained on types 3 and 4 performed better on types 2 and 5 than 1 and 6. Finally, the model trained on types 5 and 6 performed better on types 3 and 4 than types 1 and 2.

\section{Evaluating Individual Typology Angle against Fitzpatrick Skin Type Labels}

\subsection{Methodology}

An alternative approach to annotating images with Fitzpatrick labels is estimating skin tone via individual typology angle (ITA), which is calculated based on statistical features of image pixels and is negatively correlated with the melanin index~\cite{10.1001/jamadermatol.2015.0351}. Ideally, ITA is calculated over pixels in a segmented region highlighting only non-diseased skin~\cite{kinyanjui_estimating_2019}. But, segmentation masks are expensive to obtain, and instead of directly segmenting healthy skin, we apply the YCbCr algorithm to mask skin pixels \cite{Kolkur_2017}. We compare Fitzpatrick labels on the entire dataset with ITA calculated on the full images and the YCbCr masks. 

The YCbCr algorithm takes as input an image in RGBA color space and applies the following masking thresholds.  
\begin{align}
    R &> 95 \\ R &> G \\ R &> B \\ G &> 40 \\ B &> 20 \\ |R - G| &> 15 \\ A &> 15
\end{align} 
Then, the image is converted from RGBA to YCbCr color space, and applies a further masking along the following thresholds: 
\begin{align}
    Cr &> 135 \\ Cr &\geq (0.3448\cdot Cb)+76.2069 \\ Cr &\geq (-4.5652\cdot Cb)+234.5652 \\ Cr &\leq (-1.15\cdot Cb)+301.75 \\ Cr &\leq (-2.2857\cdot Cb)+432.85
\end{align} 
where $R-G-B-A$ are the respective Red-Green-Blue-Alpha components of the input image, and $Y-Cb-Cr$ are the respective luminance and chrominance components of the color-converted image. As a result, the YCbCr algorithm attempts to segment healthy skin from the rest of an image.

We calculate the ITA of each full and YCbCr masked image by converting the input image to $CIE-LAB$ color space, which contains $L$: luminance and $B$: yellow, and applying the following formula \cite{merler2019diversity}: 
\begin{equation}
    ITA = arctan(\dfrac{L^* - 50}{B^*}) \cdot \dfrac{180}{\pi}
\end{equation}
where $L^*$ and $B^*$ are the mean of non-masked pixels with values within one standard deviation of the actual mean.

\subsection{Results}

In Table~\ref{fig:T3}, we compare ITA calculations on both the full images and YCbCr masks with Fitzpatrick skin type labels. Furthermore, we compare two different methods for calculating Fitzpatrick type given ITA, as described in Equations \ref{eq:14} and \ref{eq:15}. For each entry, we calculate the proportion of ITA scores in the range of plus or minus one of the annotated Fitzpatrick score.

\begin{equation}
    Fitzpatrick(ITA) = 
    \begin{cases} 
        1 & ITA > 55 \\
        2 & 55 \geq ITA > 41 \\
        3 & 41 \geq ITA > 28 \\
        4 & 28 \geq ITA > 19 \\
        5 & 19 \geq ITA > 10 \\
        6 & 10 \geq ITA \\
    \end{cases}
\label{eq:14}
\end{equation}

\begin{equation}
    Fitzpatrick(ITA) = 
    \begin{cases} 
        1 & ITA > 40 \\
        2 & 40 \geq ITA > 23 \\
        3 & 23 \geq ITA > 12 \\
        4 & 12 \geq ITA > 0 \\
        5 & 0 \geq ITA > -25 \\
        6 & -25 \geq ITA \\
    \end{cases}
    \label{eq:15}
\end{equation}

\begin{figure*}
\begin{center}
\includegraphics[width=0.9\linewidth]{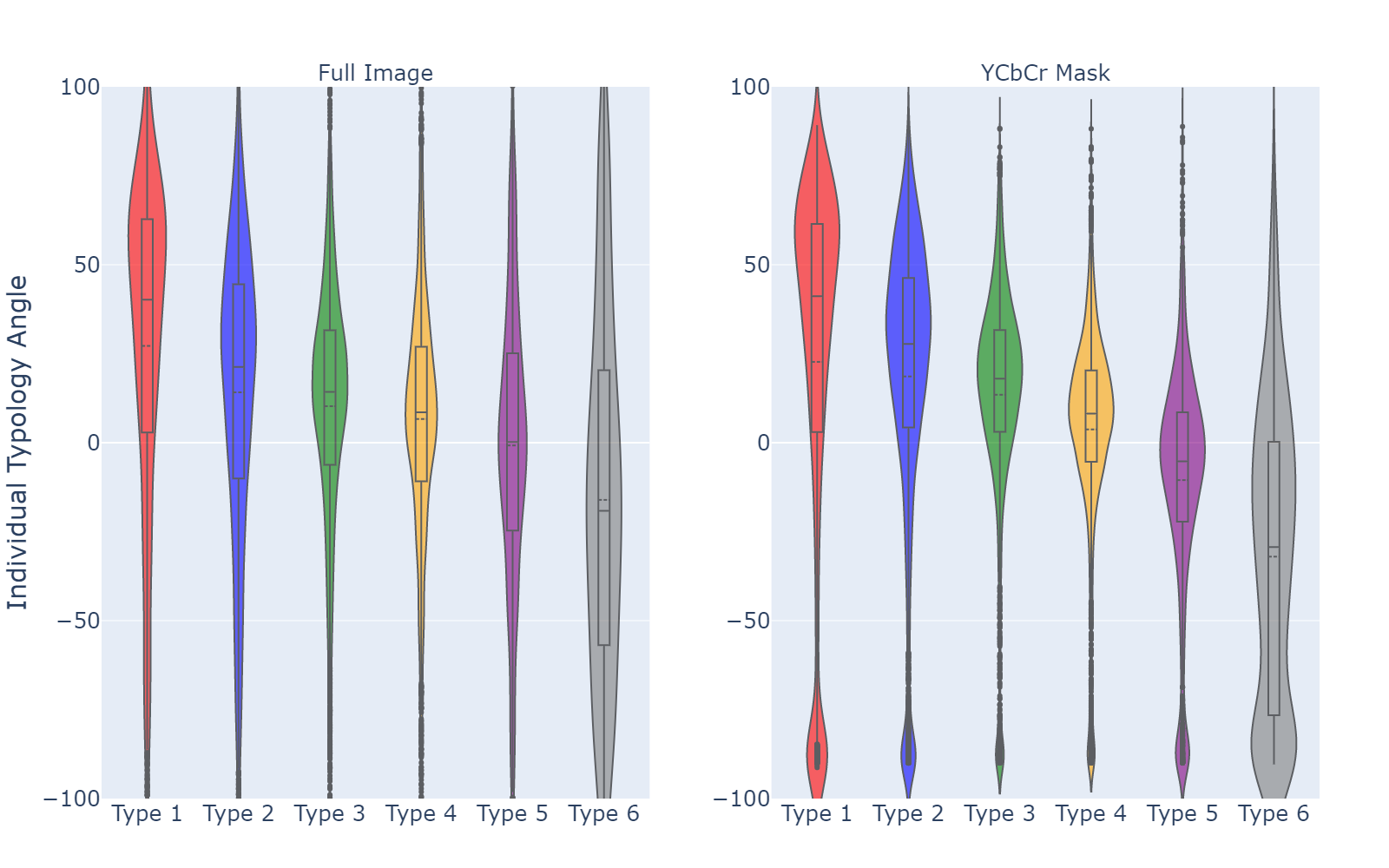}
\end{center}
   \caption{Observed distribution of individual typology angles by Fitzpatrick.}
\label{fig:violin}
\end{figure*}

\begin{figure*}[ht]
    \centering
    \includegraphics[width=0.98\textwidth]{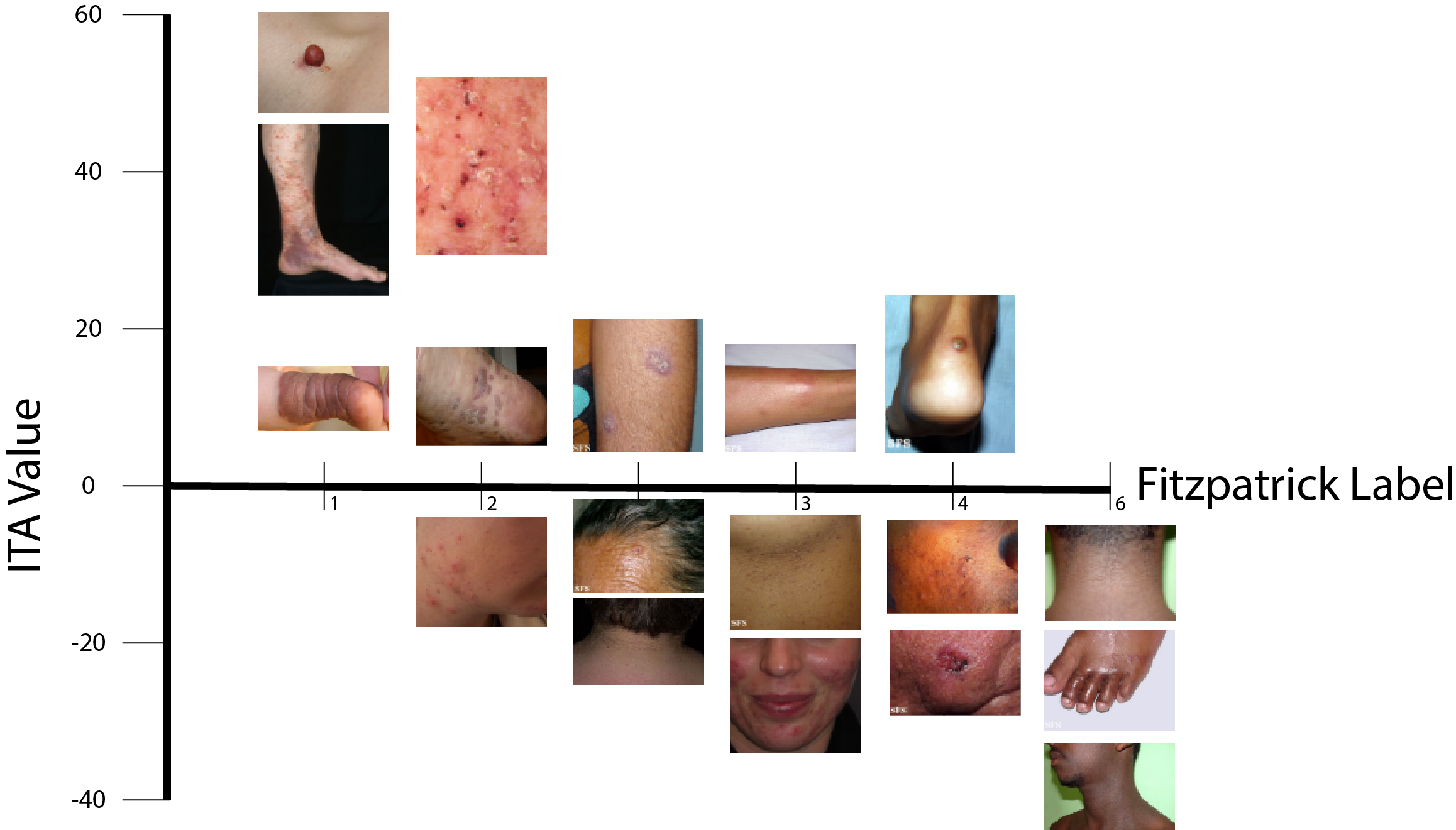}
    \caption{18 images plot arranged based on ITA values and Fitzpatrick labels.}
    \label{fig:itavfitz}
\end{figure*}

In Table~\ref{fig:T3}, the columns labeled ``Kinyananjui'' compare Fitzpatrick skin type labels with ITA following Equation \ref{eq:14}, a modified version of the ITA thresholds described by Kinyanjui et al. \cite{kinyanjui_estimating_2019}. The columns labeled ``Empirical'' follow Equation \ref{eq:15}, which we developed based on the empirical distribution of ITA scores minimizing overall error. Figure~\ref{fig:violin} plots the empirical distribution of ITA scores for each Fitzpatrick skin type label. The discrepancy between Fitzpatrick skin type labels and the ITA approach appears to be driven mostly by high variance in the ITA algorithm as Figure \ref{fig:itavfitz} reveals.

\begin{table}[ht]
\centering 
\begin{tabular}{c | cc | cc }
\toprule
    & \multicolumn{2}{|c|}{Full Image} &
\multicolumn{2}{|c}{YCbCr Mask} \\ \cline{2-5}
 & Kinyanjui & Empirical & Kinyanjui & Empirical  \\
\midrule
Overall              & 45.87\% & 60.34\%  & 53.30\%  & 70.38\%  \\
Type 1               & 50.97\% & 65.35\%  & 52.22\%  & 66.00\%  \\
Type 2               & 42.60\% & 59.57\%  & 49.15\%  & 69.47\%  \\
Type 3               & 35.43\% & 55.20\%  & 45.13\%  & 66.41\%  \\
Type 4               & 34.09\% & 58.54\%  & 40.24\%  & 72.10\%  \\
Type 5               & 78.21\% & 65.49\%  & 93.41\%  & 82.26\%  \\
Type 6               & 74.80\% & 65.04\%  & 90.71\%  & 79.69\%  \\ 
\bottomrule
\end{tabular}
\caption{Plus or minus one concordance of individual typology angle (ITA) with Fitzpatrick skin type labels. Each column shows the percent of ITA scores that are within plus or minus 1 point of the annotated Fitzpatrick labels after converting ITA to Fitzpatrick types via Equations~\ref{eq:14} and~\ref{eq:15}.}
\label{fig:T3}
\end{table}



\begin{figure*}[ht]
    \centering
    \includegraphics[width=0.32\textwidth]{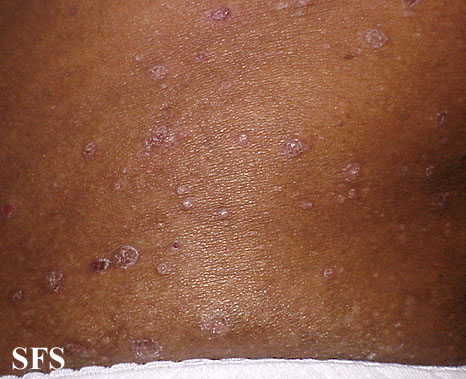}
   \includegraphics[width=0.32\textwidth]{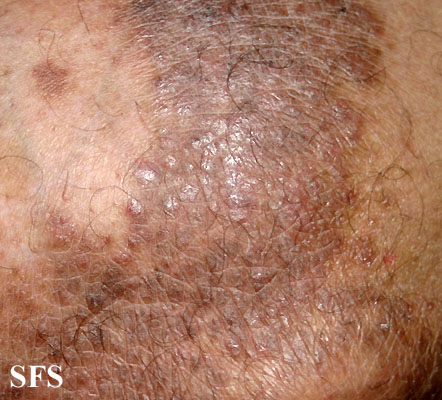}
    \includegraphics[width=0.32\textwidth]{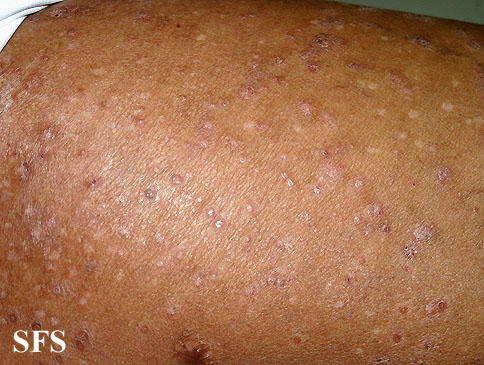}
    \includegraphics[width=0.32\textwidth]{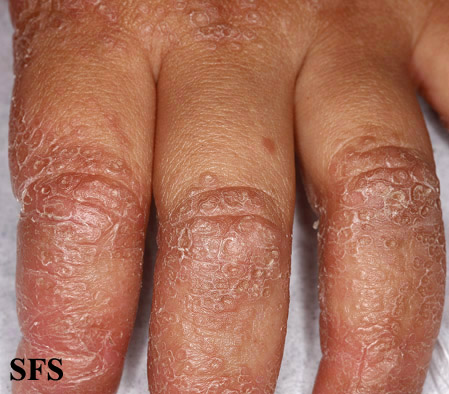}
    \includegraphics[width=0.32\textwidth]{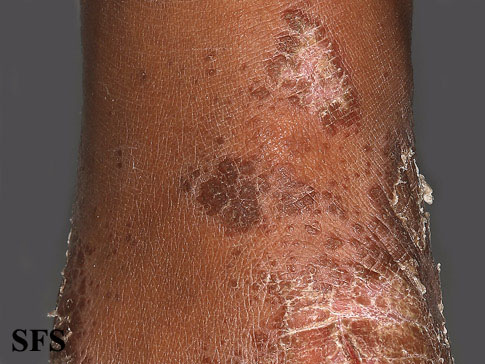}
    \includegraphics[width=0.32\textwidth]{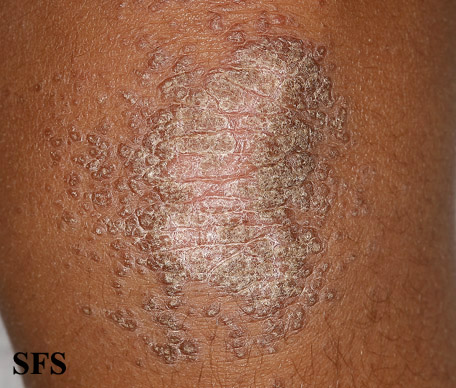}
    \caption{Example images of pityriasis rubra pilaris from Atlas Dermatologico that were accurately classified by the neural network trained on DermaAmin images. On the 174 images from Atlas Dermatologico labeled pityriasis rubra pilaris, 24\% are accurately identified, 35\% are accurately identified in the top 2 most likely predictions, and 45\% are accurately identified in the top 3 most likely predictions.}
    \label{fig:exampleimages}
\end{figure*}

\section{Conclusion}

We present the \textit{Fitzpatrick 17k}, a new dataset consisting of 16,577 clinical images of 114 different skin conditions annotated with Fitzpatrick skin type labels. These images are sourced from Atlas Dermatologico and Derma Amin and contain 3.6 times more images of the two lightest Fitzpatrick skin types than the two darkest Fitzpatrick skin types. By annotating this dataset with Fitzpatrick skin type labels, we reveal both an underrepresentation of dark skin images in online dermatology atlases and accuracy disparities that arise from training a neural network on only a subset of skin types.

By training a deep neural network based on an adapted VGG-16 architecture pre-trained on ImageNet, we achieve accuracy results that approach the levels reported on a much larger dataset~\cite{esteva_dermatologist-level_2017}. We find that the skin type in the images on which a model is trained affects the accuracy scores across Fitzpatrick skin types. Specifically, we find that models trained on data from only two Fitzpatrick skin types are most accurate on holdout images of the closest Fitzpatrick skin types to the training data. These relationships between the type of training data and holdout accuracy across skin types are consistent with what has been long known by dermatologists: skin conditions appear differently across skin types~\cite{adelekun2021skin}. 

An open question for future research is in which skin conditions do accuracy disparities appear largest across skin types. Recent research shows that diagnoses by medical students and physicians appears to vary across skin types~\cite{fenton2020medical, diao2021representation}. Future research at the intersection of dermatology and computer vision should focus on specific groups of skin conditions where accuracy disparities are expected to arise because visual features of skin conditions (e.g. redness in inflammatory conditions) do not appear universally across skin types.

The large set of Fitzpatrick skin type labels enable an empirical evaluation of ITA as an automated tool for assessing skin tone. Our comparison reveals that ITA is prone to error on images that human labelers can easily agree upon. The most accurate ITA scores are off by more than one point on the Fitzpatrick scale in about one third of the dataset. One limitation of this comparison is that we calculated ITA based on either the entire image or an automatic segmentation mask. Future work should refine this comparison based on more precise segmentation masks.

We present this dataset and paper in the hopes that it inspires future research at the intersection of dermatology and computer vision to evaluate accuracy across sub-populations where classification accuracy is suspected to be heterogeneous.

{\small
\bibliographystyle{ieee_fullname}
\bibliography{egbib}

\begin{thebibliography}{10}\itemsep=-1pt

\bibitem{abbasi-sureshjani_risk_2020}
Samaneh Abbasi-Sureshjani, Ralf Raumanns, Britt E.~J. Michels, Gerard Schouten,
  and Veronika Cheplygina.
\newblock Risk of {Training} {Diagnostic} {Algorithms} on {Data} with
  {Demographic} {Bias}.
\newblock {\em arXiv:2005.10050 [cs, stat]}, June 2020.
\newblock arXiv: 2005.10050.

\bibitem{adamson_machine_2018}
Adewole~S. Adamson and Avery Smith.
\newblock Machine {Learning} and {Health} {Care} {Disparities} in
  {Dermatology}.
\newblock {\em JAMA Dermatology}, 154(11):1247, Nov. 2018.

\bibitem{adelekun2021skin}
Ademide Adelekun, Ginikanwa Onyekaba, and Jules~B Lipoff.
\newblock Skin color in dermatology textbooks: an updated evaluation and
  analysis.
\newblock {\em Journal of the American Academy of Dermatology}, 84(1):194--196,
  2021.

\bibitem{alkattash}
Jehad~Amin AlKattash.
\newblock Dermaamin.
\newblock {\em https://www.dermaamin.com/site/}.

\bibitem{alvarado_representation_2020}
Savannah~M. Alvarado and Hao Feng.
\newblock Representation of dark skin images of common dermatologic conditions
  in educational resources: a cross-sectional analysis.
\newblock {\em Journal of the American Academy of Dermatology}, page
  S0190962220311385, June 2020.

\bibitem{arifin2012dermatological}
M~Shamsul Arifin, M~Golam Kibria, Adnan Firoze, M~Ashraful Amini, and Hong Yan.
\newblock Dermatological disease diagnosis using color-skin images.
\newblock In {\em 2012 international conference on machine learning and
  cybernetics}, volume~5, pages 1675--1680. IEEE, 2012.

\bibitem{barata2021explainable}
Catarina Barata, M~Emre Celebi, and Jorge~S Marques.
\newblock Explainable skin lesion diagnosis using taxonomies.
\newblock {\em Pattern Recognition}, 110:107413, 2021.

\bibitem{barocas2016big}
Solon Barocas and Andrew~D Selbst.
\newblock Big data's disparate impact.
\newblock {\em Calif. L. Rev.}, 104:671, 2016.

\bibitem{bissoto2018deep}
Alceu Bissoto, F{\'a}bio Perez, Vin{\'\i}cius Ribeiro, Michel Fornaciali,
  Sandra Avila, and Eduardo Valle.
\newblock Deep-learning ensembles for skin-lesion segmentation, analysis,
  classification: Recod titans at isic challenge 2018.
\newblock {\em arXiv preprint arXiv:1808.08480}, 2018.

\bibitem{bissoto2020debiasing}
Alceu Bissoto, Eduardo Valle, and Sandra Avila.
\newblock Debiasing skin lesion datasets and models? not so fast.
\newblock In {\em Proceedings of the IEEE/CVF Conference on Computer Vision and
  Pattern Recognition Workshops}, pages 740--741, 2020.

\bibitem{buolamwini2018gender}
Joy Buolamwini and Timnit Gebru.
\newblock Gender shades: Intersectional accuracy disparities in commercial
  gender classification.
\newblock In {\em Conference on fairness, accountability and transparency},
  pages 77--91. PMLR, 2018.

\bibitem{celebi2019dermoscopy}
M~Emre Celebi, Noel Codella, and Allan Halpern.
\newblock Dermoscopy image analysis: overview and future directions.
\newblock {\em IEEE journal of biomedical and health informatics},
  23(2):474--478, 2019.

\bibitem{chen2018my}
Irene Chen, Fredrik~D Johansson, and David Sontag.
\newblock Why is my classifier discriminatory?
\newblock {\em arXiv preprint arXiv:1805.12002}, 2018.

\bibitem{chen2020ethical}
Irene~Y Chen, Emma Pierson, Sherri Rose, Shalmali Joshi, Kadija Ferryman, and
  Marzyeh Ghassemi.
\newblock Ethical machine learning in health.
\newblock {\em arXiv preprint arXiv:2009.10576}, 2020.

\bibitem{codella2019skin}
Noel Codella, Veronica Rotemberg, Philipp Tschandl, M~Emre Celebi, Stephen
  Dusza, David Gutman, Brian Helba, Aadi Kalloo, Konstantinos Liopyris, Michael
  Marchetti, et~al.
\newblock Skin lesion analysis toward melanoma detection 2018: A challenge
  hosted by the international skin imaging collaboration (isic).
\newblock {\em arXiv preprint arXiv:1902.03368}, 2019.

\bibitem{codella2018skin}
Noel~CF Codella, David Gutman, M~Emre Celebi, Brian Helba, Michael~A Marchetti,
  Stephen~W Dusza, Aadi Kalloo, Konstantinos Liopyris, Nabin Mishra, Harald
  Kittler, et~al.
\newblock Skin lesion analysis toward melanoma detection: A challenge at the
  2017 international symposium on biomedical imaging (isbi), hosted by the
  international skin imaging collaboration (isic).
\newblock In {\em 2018 IEEE 15th International Symposium on Biomedical Imaging
  (ISBI 2018)}, pages 168--172. IEEE, 2018.

\bibitem{international2020siim}
International Skin~Imaging Collaboration et~al.
\newblock Siim-isic 2020 challenge dataset.
\newblock {\em International Skin Imaging Collaboration}, 2020.

\bibitem{combalia2019bcn20000}
Marc Combalia, Noel~CF Codella, Veronica Rotemberg, Brian Helba, Veronica
  Vilaplana, Ofer Reiter, Cristina Carrera, Alicia Barreiro, Allan~C Halpern,
  Susana Puig, et~al.
\newblock Bcn20000: Dermoscopic lesions in the wild.
\newblock {\em arXiv preprint arXiv:1908.02288}, 2019.

\bibitem{cowgill2020algorithmic}
Bo Cowgill and Catherine~E Tucker.
\newblock Algorithmic fairness and economics.
\newblock {\em The Journal of Economic Perspectives}, 2020.

\bibitem{deng2009imagenet}
Jia Deng, Wei Dong, Richard Socher, Li-Jia Li, Kai Li, and Li Fei-Fei.
\newblock Imagenet: A large-scale hierarchical image database.
\newblock In {\em 2009 IEEE conference on computer vision and pattern
  recognition}, pages 248--255. Ieee, 2009.

\bibitem{diao2021representation}
James~A Diao and Adewole~S Adamson.
\newblock Representation and misdiagnosis of dark skin in a large-scale visual
  diagnostic challenge.
\newblock {\em Journal of the American Academy of Dermatology}, 2021.

\bibitem{dulmage_point--care_2020}
Brittany Dulmage, Kyle Tegtmeyer, Michael~Z. Zhang, Maria Colavincenzo, and
  Shuai Xu.
\newblock A {Point}-of-{Care}, {Real}-{Time} {Artificial} {Intelligence}
  {System} to {Support} {Clinician} {Diagnosis} of a {Wide} {Range} of {Skin}
  {Diseases}.
\newblock {\em Journal of Investigative Dermatology}, page S0022202X20321679,
  Oct. 2020.

\bibitem{esteva_dermatologist-level_2017}
Andre Esteva, Brett Kuprel, Roberto~A. Novoa, Justin Ko, Susan~M. Swetter,
  Helen~M. Blau, and Sebastian Thrun.
\newblock Dermatologist-level classification of skin cancer with deep neural
  networks.
\newblock {\em Nature}, 542(7639):115--118, Feb. 2017.

\bibitem{fenton2020medical}
Anne Fenton, Erika Elliott, Ashkan Shahbandi, Ekene Ezenwa, Chance Morris,
  Justin McLawhorn, James~G Jackson, Pamela Allen, and Andrea Murina.
\newblock Medical students’ ability to diagnose common dermatologic
  conditions in skin of color.
\newblock {\em Journal of the American Academy of Dermatology}, 83(3):957,
  2020.

\bibitem{fitzpatrick1988validity}
Thomas~B Fitzpatrick.
\newblock The validity and practicality of sun-reactive skin types i through
  vi.
\newblock {\em Archives of dermatology}, 124(6):869--871, 1988.

\bibitem{desilva}
Samuel Freire~da Silva.
\newblock Atlas dermatologico.
\newblock {\em http://atlasdermatologico.com.br/}.

\bibitem{giotis2015med}
Ioannis Giotis, Nynke Molders, Sander Land, Michael Biehl, Marcel~F Jonkman,
  and Nicolai Petkov.
\newblock Med-node: A computer-assisted melanoma diagnosis system using
  non-dermoscopic images.
\newblock {\em Expert systems with applications}, 42(19):6578--6585, 2015.

\bibitem{gupta_skin_nodate}
Alpana~K Gupta, Mausumi Bharadwaj, and Ravi Mehrotra.
\newblock Skin cancer concerns in people of color: risk factors and prevention.
\newblock {\em Asian Pacific journal of cancer prevention: APJCP}, 17(12):5257,
  2016.

\bibitem{han2018classification}
Seung~Seog Han, Myoung~Shin Kim, Woohyung Lim, Gyeong~Hun Park, Ilwoo Park, and
  Sung~Eun Chang.
\newblock Classification of the clinical images for benign and malignant
  cutaneous tumors using a deep learning algorithm.
\newblock {\em Journal of Investigative Dermatology}, 138(7):1529--1538, 2018.

\bibitem{kawahara2018seven}
Jeremy Kawahara, Sara Daneshvar, Giuseppe Argenziano, and Ghassan Hamarneh.
\newblock Seven-point checklist and skin lesion classification using multitask
  multimodal neural nets.
\newblock {\em IEEE journal of biomedical and health informatics},
  23(2):538--546, 2018.

\bibitem{kinyanjui_estimating_2019}
Newton~M. Kinyanjui, Timothy Odonga, Celia Cintas, Noel C.~F. Codella, Rameswar
  Panda, Prasanna Sattigeri, and Kush~R. Varshney.
\newblock Estimating {Skin} {Tone} and {Effects} on {Classification}
  {Performance} in {Dermatology} {Datasets}.
\newblock {\em arXiv:1910.13268 [cs, stat]}, Oct. 2019.
\newblock arXiv: 1910.13268.

\bibitem{kleinberg_algorithms_2020}
Jon Kleinberg, Jens Ludwig, Sendhil Mullainathan, and Cass~R. Sunstein.
\newblock Algorithms as discrimination detectors.
\newblock {\em Proceedings of the National Academy of Sciences}, page
  201912790, July 2020.

\bibitem{Kolkur_2017}
S. Kolkur, D. Kalbande, P. Shimpi, C. Bapat, and J. Jatakia.
\newblock Human skin detection using rgb, hsv and ycbcr color models.
\newblock {\em Proceedings of the International Conference on Communication and
  Signal Processing 2016 (ICCASP 2016)}, 2017.

\bibitem{lester_absence_2020}
J.C. Lester, J.L. Jia, L. Zhang, G.A. Okoye, and E. Linos.
\newblock Absence of images of skin of colour in publications of {COVID}‐19
  skin manifestations.
\newblock {\em British Journal of Dermatology}, 183(3):593--595, Sept. 2020.

\bibitem{lester_diversity_nodate}
Jenna Lester and Kanade Shinkai.
\newblock Diversity and inclusivity are essential to the future of dermatology.
\newblock {\em Cutis}, 104(2):99--100, 2019.

\bibitem{liu_deep_2020}
Yuan Liu, Ayush Jain, Clara Eng, David~H. Way, Kang Lee, Peggy Bui, Kimberly
  Kanada, Guilherme de Oliveira~Marinho, Jessica Gallegos, Sara Gabriele,
  Vishakha Gupta, Nalini Singh, Vivek Natarajan, Rainer Hofmann-Wellenhof,
  Greg~S. Corrado, Lily~H. Peng, Dale~R. Webster, Dennis Ai, Susan~J. Huang,
  Yun Liu, R.~Carter Dunn, and David Coz.
\newblock A deep learning system for differential diagnosis of skin diseases.
\newblock {\em Nature Medicine}, 26(6):900--908, June 2020.

\bibitem{mendoncca2013ph}
Teresa Mendon{\c{c}}a, Pedro~M Ferreira, Jorge~S Marques, Andr{\'e}~RS Marcal,
  and Jorge Rozeira.
\newblock Ph 2-a dermoscopic image database for research and benchmarking.
\newblock In {\em 2013 35th annual international conference of the IEEE
  engineering in medicine and biology society (EMBC)}, pages 5437--5440. IEEE,
  2013.

\bibitem{merler2019diversity}
Michele Merler, Nalini Ratha, Rogerio~S. Feris, and John~R. Smith.
\newblock Diversity in faces, 2019.

\bibitem{northcutt2021pervasive}
Curtis~G Northcutt, Anish Athalye, and Jonas Mueller.
\newblock Pervasive label errors in test sets destabilize machine learning
  benchmarks.
\newblock {\em arXiv preprint arXiv:2103.14749}, 2021.

\bibitem{oakden-rayner_hidden_2020}
Luke Oakden-Rayner, Jared Dunnmon, Gustavo Carneiro, and Christopher Re.
\newblock Hidden stratification causes clinically meaningful failures in
  machine learning for medical imaging.
\newblock In {\em Proceedings of the {ACM} {Conference} on {Health},
  {Inference}, and {Learning}}, pages 151--159, Toronto Ontario Canada, Apr.
  2020. ACM.

\bibitem{obermeyer_dissecting_2019}
Ziad Obermeyer, Brian Powers, Christine Vogeli, and Sendhil Mullainathan.
\newblock Dissecting racial bias in an algorithm used to manage the health of
  populations.
\newblock {\em Science}, 366(6464):447--453, Oct. 2019.

\bibitem{pacheco2020pad}
Andre~GC Pacheco, Gustavo~R Lima, Amanda~S Salom{\~a}o, Breno Krohling, Igor~P
  Biral, Gabriel~G de Angelo, F{\'a}bio~CR Alves~Jr, Jos{\'e}~GM Esgario,
  Alana~C Simora, Pedro~BC Castro, et~al.
\newblock Pad-ufes-20: A skin lesion dataset composed of patient data and
  clinical images collected from smartphones.
\newblock {\em Data in brief}, 32:106221, 2020.

\bibitem{palmieri_missed_2019}
James~R Palmieri.
\newblock Missed {Diagnosis} and the {Development} of {Acute} and {Late} {Lyme}
  {Disease} in {Dark} {Skinned} {Populations} of {Appalachia}.
\newblock {\em Biomedical Journal of Scientific \& Technical Research}, 21(2),
  Sept. 2019.

\bibitem{pierson2021algorithmic}
Emma Pierson, David~M Cutler, Jure Leskovec, Sendhil Mullainathan, and Ziad
  Obermeyer.
\newblock An algorithmic approach to reducing unexplained pain disparities in
  underserved populations.
\newblock {\em Nature Medicine}, 27(1):136--140, 2021.

\bibitem{ramezani2014automatic}
Maryam Ramezani, Alireza Karimian, and Payman Moallem.
\newblock Automatic detection of malignant melanoma using macroscopic images.
\newblock {\em Journal of medical signals and sensors}, 4(4):281, 2014.

\bibitem{rotemberg2021patient}
Veronica Rotemberg, Nicholas Kurtansky, Brigid Betz-Stablein, Liam Caffery,
  Emmanouil Chousakos, Noel Codella, Marc Combalia, Stephen Dusza, Pascale
  Guitera, David Gutman, et~al.
\newblock A patient-centric dataset of images and metadata for identifying
  melanomas using clinical context.
\newblock {\em Scientific data}, 8(1):1--8, 2021.

\bibitem{simonyan2014very}
Karen Simonyan and Andrew Zisserman.
\newblock Very deep convolutional networks for large-scale image recognition.
\newblock {\em arXiv preprint arXiv:1409.1556}, 2014.

\bibitem{soenksen2021using}
Luis~R Soenksen, Timothy Kassis, Susan~T Conover, Berta Marti-Fuster, Judith~S
  Birkenfeld, Jason Tucker-Schwartz, Asif Naseem, Robert~R Stavert, Caroline~C
  Kim, Maryanne~M Senna, et~al.
\newblock Using deep learning for dermatologist-level detection of suspicious
  pigmented skin lesions from wide-field images.
\newblock {\em Science Translational Medicine}, 13(581), 2021.

\bibitem{sun2016benchmark}
Xiaoxiao Sun, Jufeng Yang, Ming Sun, and Kai Wang.
\newblock A benchmark for automatic visual classification of clinical skin
  disease images.
\newblock In {\em European Conference on Computer Vision}, pages 206--222.
  Springer, 2016.

\bibitem{swinney2015incontinentia}
Christian~C Swinney, Dennis~P Han, and Peter~A Karth.
\newblock Incontinentia pigmenti: a comprehensive review and update.
\newblock {\em Ophthalmic Surgery, Lasers and Imaging Retina}, 46(6):650--657,
  2015.

\bibitem{tschandl_humancomputer_2020}
Philipp Tschandl, Christoph Rinner, Zoe Apalla, Giuseppe Argenziano, Noel
  Codella, Allan Halpern, Monika Janda, Aimilios Lallas, Caterina Longo, Josep
  Malvehy, John Paoli, Susana Puig, Cliff Rosendahl, H.~Peter Soyer, Iris
  Zalaudek, and Harald Kittler.
\newblock Human–computer collaboration for skin cancer recognition.
\newblock {\em Nature Medicine}, 26(8):1229--1234, Aug. 2020.

\bibitem{tschandl2018ham10000}
Philipp Tschandl, Cliff Rosendahl, and Harald Kittler.
\newblock The ham10000 dataset, a large collection of multi-source
  dermatoscopic images of common pigmented skin lesions.
\newblock {\em Scientific data}, 5(1):1--9, 2018.

\bibitem{van2015cutaneous}
Lindi Van~Zyl, Jeanetta Du~Plessis, and Joe Viljoen.
\newblock Cutaneous tuberculosis overview and current treatment regimens.
\newblock {\em Tuberculosis}, 95(6):629--638, 2015.

\bibitem{ware2020racial}
Olivia~R Ware, Jessica~E Dawson, Michi~M Shinohara, and Susan~C Taylor.
\newblock Racial limitations of fitzpatrick skin type.
\newblock {\em Cutis.}, 105(2):77--80, 2020.

\bibitem{10.1001/jamadermatol.2015.0351}
Marcus Wilkes, Caradee~Y. Wright, Johan~L. du Plessis, and Anthony Reeder.
\newblock {Fitzpatrick Skin Type, Individual Typology Angle, and Melanin Index
  in an African Population: Steps Toward Universally Applicable Skin
  Photosensitivity Assessments}.
\newblock {\em JAMA Dermatology}, 151(8):902--903, 08 2015.

\bibitem{winkler_association_2019}
Julia~K. Winkler, Christine Fink, Ferdinand Toberer, Alexander Enk, Teresa
  Deinlein, Rainer Hofmann-Wellenhof, Luc Thomas, Aimilios Lallas, Andreas
  Blum, Wilhelm Stolz, and Holger~A. Haenssle.
\newblock Association {Between} {Surgical} {Skin} {Markings} in {Dermoscopic}
  {Images} and {Diagnostic} {Performance} of a {Deep} {Learning}
  {Convolutional} {Neural} {Network} for {Melanoma} {Recognition}.
\newblock {\em JAMA Dermatology}, 155(10):1135, Oct. 2019.

\end{thebibliography}
}

\end{document}